\pdfoutput=1
\documentclass[11pt,a4paper]{article}
\usepackage[hyperref]{acl2021}
\usepackage{times}
\usepackage{latexsym}
\usepackage{caption, subcaption}
\usepackage{graphicx}

\usepackage{bbm}
\usepackage{amsmath}
\usepackage{amsfonts}
\usepackage{amssymb}

\usepackage{tikz,pgfplots}
\usepackage{tikz-dependency}

\usepackage{multicol, multirow}
\usepackage{hhline}
\usepackage{boldline}
\usepackage{mathtools}
\usepackage{enumitem}
\usepackage{hyperref}

\usepackage{amsmath,amsfonts,bm}

\def\eqref#1{equation~\ref{#1}}

\def\1{\bm{1}}

\def\rw{{\textnormal{w}}}

\def\vtheta{{\bm{\theta}}}
\def\va{{\bm{a}}}

\def\vr{{\bm{r}}}

\def\vv{{\bm{v}}}
\def\vw{{\bm{w}}}
\def\vx{{\bm{x}}}

\def\eva{{a}}

\def\evr{{r}}

\def\mR{{\bm{R}}}

\def\mV{{\bm{V}}}

\def\mY{{\bm{Y}}}

\DeclareMathAlphabet{\mathsfit}{\encodingdefault}{\sfdefault}{m}{sl}
\SetMathAlphabet{\mathsfit}{bold}{\encodingdefault}{\sfdefault}{bx}{n}

\newcommand{\E}{\mathbb{E}}

\newcommand{\R}{\mathbb{R}}

\newcommand{\softmax}{\mathrm{softmax}}

\aclfinalcopy 

\title{Enhanced Universal Dependency Parsing with Automated Concatenation of Embeddings}

\author{Xinyu Wang$^{\diamond\spadesuit}$, Zixia Jia$^{\diamond\spadesuit}$, Yong Jiang$^{\dagger}$,  Kewei Tu$^{\diamond}$\thanks{$^{\spadesuit}$: Equal contributions.} \\
 $^\diamond$School of Information Science and Technology, ShanghaiTech University \\
 Shanghai Engineering Research Center of Intelligent Vision and Imaging \\
 $^\dagger$DAMO Academy, Alibaba Group \\
  {\tt \{wangxy1,jiazx,tukw\}@shanghaitech.edu.cn} \\
  {\tt \{yongjiang.jy\}@alibaba-inc.com} \\
}

\date{}

\begin{document}
\maketitle
\begin{abstract}
This paper describes the system used in submission from SHANGHAITECH team to the \textit{IWPT 2021 Shared Task}. Our system is a graph-based parser with the technique of Automated Concatenation of Embeddings (ACE). Because recent work found that better word representations can be obtained by concatenating different types of embeddings, we use ACE to automatically find the better concatenation of embeddings for the task of enhanced universal dependencies. According to official results averaged on 17 languages, our system ranks 2nd over 9 teams.
\end{abstract}

\section{Introduction}
Compared to the Universal Dependencies (UD) \cite{nivre-etal-2016-universal}, the Enhanced Universal Dependencies (EUD) \citep{EUDparsingST:2020,bouma-etal-2021-overview}\footnote{\url{https://universaldependencies.org/u/overview/enhanced-syntax.html}} makes some of the implicit relations between words more explicit and augments some of the dependency labels to facilitate the disambiguation of types of arguments and modifiers. The representation of EUD is an enhanced graph with reentrancies, cycles, and empty nodes. Such representation can represent richer grammatical relations than rooted trees, but it is harder to learn. To make the learning process relatively easy, we transfer the enhanced graph to a bi-lexical structure like annotation of semantic dependency parsing (SDP) \cite{oepen2015semeval} by reducing reentrancies and empty nodes into new labels. Therefore, many approaches for SDP can be adopted by EUD. Instead of the second-order parser that was used in previous work \cite{wang-etal-2019-second,wang-etal-2020-enhanced,wang-tu-2020-second}, we apply the biaffine parser \citep{dozat2018simpler} which is one of the state-of-the-art approaches of SDP for simplicity.

Recent developments on pre-trained contextualized embeddings have significantly improved the performance of structured prediction tasks in natural language processing. A lot of work has also shown that word representations based on the concatenation of multiple pre-trained contextualized embeddings and traditional non-contextualized embeddings (such as word2vec \citep{mikolov2013distributed} and character embeddings \citep{santos2014learning}) can further improve performance \citep{peters-etal-2018-deep,akbik-etal-2018-contextual,strakova-etal-2019-neural,wang-etal-2020-more}. \citet{wang2020automated} proposed Automated Concatenation of Embeddings to automate the process of finding better concatenations of embeddings and further improved performance of many tasks. We utilize their method to find concatenations of pre-trained embeddings as the input of the biaffine parser for EUD. Because there are many contextualized embeddings, such as XLMR \cite{conneau2019unsupervised}, BERT \cite{devlin2018bert} and Flair \cite{akbik-etal-2018-contextual}, non-contextualized embeddings, such as  word2vec \citep{mikolov2013distributed}, GloVe \citep{pennington2014glove}, and fastText \citep{bojanowski2017enriching}, and character embeddings \citep{santos2014learning}. The search space of embeddings concatenation is large in size, besides, we need to train models of 17 languages respectively. Following \citet{wang2020automated}, we use reinforcement learning to efficiently find the better embeddings concatenation for each language. Experimental results averaged on 17 languages show the effectiveness of our approach. Our system is ranked 2nd over 9 teams in the official evaluation.

\section{System Description}
\subsection{Data Pre-processing}
We adopt the same data pre-processing method as \citet{wang-etal-2020-enhanced} which transfers EUD graphs to SDP graphs. For the reentrancies of the same head and dependent on different labels in the EUD graph, we combined these arcs into one and concatenate the labels of these arcs with a special symbol `+` representing the combination of two arcs. For the empty nodes in the EUD graph, there is an official script that can reduce such empty nodes into non-empty nodes with new dependency labels\footnote{For more details, please refer to \url{https://universaldependencies.org/iwpt20/task_and_evaluation.html}.}.

\begin{table*}[h!]
\tiny
\centering
\begin{tabular}{l|l|l}
\hline\hline
\bf \textsc{Embedding (language)} & \bf \textsc{Resource} & \bf \textsc{URL}\\
\hline
fastText (all) & \citet{bojanowski2017enriching} & \url{github.com/facebookresearch/fastText}\\
M-BERT (all) & \citet{devlin-etal-2019-bert} & \url{huggingface.co/bert-base-multilingual-cased}\\
BERT (en, et, sk, ta, uk) & \citet{devlin-etal-2019-bert} & \url{huggingface.co/bert-base-cased}\\
BERT (ar) & \citet{safaya-etal-2020-kuisail} & \url{huggingface.co/asafaya/bert-large-arabic}\\
BERT (bg, cs, pl, ru) & \citet{arkhipov-etal-2019-tuning} & \url{huggingface.co/DeepPavlov/bert-base-bg-cs-pl-ru-cased}\\
BERT (fi) & \citet{DBLP:journals/corr/abs-1912-07076} & \url{huggingface.co/TurkuNLP/bert-base-finnish-cased-v1}\\
BERT (fr) & \citet{martin-etal-2020-camembert} & \url{huggingface.co/camembert-base}\\
BERT (it) & dbmdz & \url{huggingface.co/dbmdz/bert-base-italian-cased}\\
BERT (lt) & U\&R\shortcite{20.500.11821/42} & \url{huggingface.co/EMBEDDIA/litlat-bert}\\
BERT (lv) & U\&R\shortcite{20.500.11821/42} & \url{huggingface.co/EMBEDDIA/litlat-bert}\\
BERT (nl) & wietsedv & \url{huggingface.co/wietsedv/bert-base-dutch-cased}\\
BERT (sv) & \citet{swedish-bert} & \url{huggingface.co/KB/bert-base-swedish-cased}\\
XLM-R (all) & \citet{conneau-etal-2020-unsupervised} & \url{huggingface.co/xlm-roberta-large}\\
RoBERTa (uk) & youscan & \url{huggingface.co/youscan/ukr-roberta-base}\\
RoBERTa (ru) & \citet{blinov-avetisian-2020-transformer} & \url{huggingface.co/blinoff/roberta-base-russian-v0}\\
RoBERTa (nl) & \citet{delobelle-etal-2020-robbert} & \url{huggingface.co/pdelobelle/robbert-v2-dutch-base}\\
RoBERTa (others) & \citet{liu2019roberta} & \url{huggingface.co/roberta-large}\\
XLNet (en) & \citet{yang2019xlnet} & \url{huggingface.co/xlnet-large-cased}\\
\hline\hline
\end{tabular}
\caption{The embeddings we used in our system. The URL is where we downloaded the embeddings. `all' means we use the model for all the languages. `other' means we use this RoBERTa model for all the languages except the uk, ru and nl.}
\label{tab:embeddings}
\end{table*}

\subsection{Approach}
We follow the approach of \citet{wang2020automated}\footnote{\url{https://github.com/Alibaba-NLP/ACE}. Our code will be released here as well.} to build our system. Our system contains two parts: an ACE module to determine embedding concatenation as inputs, a biaffine parser to predict edges' existence and labels between each word pair. We introduce these two parts respectively.

\paragraph{ACE} Given a sentence with $n$ words $\mathbf{w}=[w_1,w_2,...,w_n]$, we first get the input representations $\mV = [\vv_1; \cdots;\vv_i;\cdots; \vv_n]$, $\mV \in \R^{d\times n}$ for the sentence, where $\vv_i$ is word representation of $i$-th word and it is a concatenation of $L$ types of word embeddings:
\begin{align}
\vv_i^l &= \text{embed}_i^l (\vx); \;\; \vv_i = [\vv_i^1;\vv_i^2; \dots; \vv_i^L] \nonumber
\end{align}
where $\text{embed}^l$ is the model of $l$-th embeddings, $\vv_i\in \R^d$, $\vv_i^l\in \R^{d^l}$. $d^l$ is the hidden size of $\text{embed}^l$. Our ACE use a binary vector $\va=[a_1, \cdots, a_l, \cdots, a_L]$ as an mask to choose a subset of embeddings of $L$ types and mask out the rest. Thus, the embeddings become:
\begin{align}
\vv_i = [\vv_i^1a_1;\dots;\vv_i^la_l; \dots; \vv_i^La_L] \label{eq:vector} \nonumber
\end{align}
where $a_l$ is a binary variable.

To learn this mask (i.e., embeddings concatenation), we set a controller which interact with our EUD parser to iteratively generate the embedding mask from the search space. Defined the probability distribution of selecting an concatenation $\va$ as $P^{\text{ctrl}}(\va;\vtheta)=\prod_{l=1}^LP_l^{\text{ctrl}}(a_l;\theta_l)$. Each element $a_l$ of $\va$ is
sampled independently from a Bernoulli distribution,
which is defined as:
\begin{align}
P_l^{\text{ctrl}}(a_l;\theta_l) {=}
    \begin{cases}
    \sigma(\theta_l) &a_l{=}1\\
    1{-} P_l^{\text{ctrl}}(a_l{=}1;\theta_l) &a_l{=}0
    \end{cases}
\end{align}
where $\sigma$ is the sigmoid function.

We use reinforcement learning and take the accuracy on development set of our EUD parser as reward signal $R$. The controller's target is to maximize the expected reward $J(\vtheta)=\E_{P^{\text{ctrl}}(\va;\vtheta)}[R]$ through the policy gradient method \citep{williams1992simple}. We defined the reward function as:
\begin{align}
\vr^t {=} \sum_{i=1}^{t-1} (R_t{-}R_i) \gamma ^{Hamm(\va^t,\va^{i})-1} |\va^t{-}\va^{i}| \label{eq:reward}
\end{align}
Where $\gamma \in (0,1)$. $|\va^t-\va^{i}|$ is a binary vector, representing the change between current embedding concatenation $\va^t$ at current time step $t$ and $\va^{i}$ at previous time step $i$. $R_t$ and $R_i$ are the reward at time step $t$ and $i$. $Hamm(\va^t,\va^{i})$ is the Hamming distance of two concatenations. 

Since calculating the exact expectation is intractable in our approach, the gradient of $J(\vtheta)$ is approximated by sampling only one selection following the distribution $P^{\text{ctrl}}(\va;\vtheta)$ at each step for training efficiency. With the reward function, the final formulation is:
\begin{align}
\nabla_\vtheta J_t(\vtheta) \approx \sum_{l=1}^L \nabla_\vtheta \log P_l^{\text{ctrl}}(\eva_l^t;\theta_l) \evr^t_{l} \label{eq:gradient}
\end{align}

\paragraph{EUD Parser} After getting the representation $\mV$ of the sentence $\vw$, we use a three-layer BiLSTM taking the representation as input:
\begin{align*}
    \mR&=\mathrm{BiLSTM}(\mV)
\end{align*}
Where $\mR=[\vr_1,\dots,\vr_n]$ represents the output from the BiLSTM. For the arc prediction and label prediction, we use two different feed-forward networks and biaffine functions:
\begin{align*}
    s_{ij}^{\mathrm{(arc)}}&=\mathrm{FNN\_Biaffine}^{\mathrm{(arc)}}(\vr_i,\vr_j) \\
    \mathbf{s}_{ij}^{\mathrm{(label)}}&=\mathrm{FNN\_Biaffine}^{\mathrm{(label)}}(\vr_i,\vr_j)
\end{align*}
The arc probability distribution and the label probability distribution for each potential arc are:
\begin{align*}
    P^{\mathrm{(arc)}}(y_{ij}^{\mathrm{(arc)}}|\rw)&=\softmax([s_{ij}^{\mathrm{(arc)}};0]) \\
    P^{\mathrm{(label)}}(y_{ij}^{\mathrm{(label)}}|\rw)&=\softmax(\mathbf{s}_{ij}^{\mathrm{(label)}})
\end{align*}
According to $s_{ij}^{\mathrm{(arc)}}$, we first use MST \cite{mcdonald2005online} algorithm to get a tree structure, then we additionally add arcs for the positions that $s_{ij}^{\mathrm{(arc)}}>0$. Such method can get a EUD graph and ensure the connectivity of the graph.  \citet{wang-etal-2020-enhanced} shows that the non-projective tree algorithm (MST) is better than the projective tree algorithm (Eisner's) for the EUD task. We select the label with the highest score of each potential arc.

Given any labeled sentence $(\vw, \mY^{\star})$, where $\mY^{\star}$ stands for a gold parse graph, to train the system, we follow the approach of \citet{wang-etal-2019-second} with the cross entropy loss:
\begin{align*}
\mathcal{L}^{\mathrm{(arc)}} (\Lambda) &= -\sum_{i,j} \log(P_\Lambda (y_{ij}^{\star\mathrm{(arc)}}|\vw))\\
\mathcal{L}^{\mathrm{(label)}} (\Lambda) &= -\sum_{i,j}\mathbbm{1}(y_{ij}^{\star\mathrm{(arc)}}) \log(P_\Lambda (y_{ij}^{\star\mathrm{(label)}}|\vw))
\end{align*}
where $\Lambda$ is the parameters of our system, $\mathbbm{1}(y_{ij}^{\star\mathrm{(arc)}})$ denotes the indicator function and equals 1 when edge $(i,j)$ exists in the gold parse and 0 otherwise, and $i,j$ ranges over all the tokens $\vw$ in the sentence.
The two losses are combined by a weighted average.
\begin{equation*}
    \mathcal{L}=\lambda\mathcal{L}^{(label)}+(1-\lambda)\mathcal{L}^{(arc)}
\end{equation*}
Where $\lambda$ is a hyper-parameter.

\begin{table}[h!]
\centering
\begin{tabular}{ccc}
\hline\hline
\multirow{2}{*}{\textbf{Language}} & \multicolumn{1}{c}{\textbf{Fine-tuned XLM-R}} & \multicolumn{1}{c}{\textbf{ACE}} \\ \cline{2-3} 
                  & \textbf{ELAS}                               & \textbf{ELAS}                    \\ \cline{2-3} 
Arabic            & 76.07                                       & 82.90                           \\
Bulgarian         & 87.92                                       & 91.46                            \\
Czech             & 91.64                                       & 92.95                            \\
Dutch             & 87.11                                       & 92.33                            \\
English           & 86.04                                       & 89.24                            \\
Estonian          & 87.13                                       & 89.37                            \\
Finnish           & 86.00                                       & 91.66                            \\
French            & 74.74                                       & 93.65                             \\
Italian           & 89.31                                       & 93.03                            \\
Latvian           & 84.83                                       & 90.11                            \\
Lithuanian        & 68.92                                       & 85.48                            \\
Polish            & 87.98                                       & 90.90                            \\
Russian           & 91.52                                       & 93.22                            \\
Slovak            & 85.26                                       & 90.92                            \\
Swedish           & 76.02                                       & 88.04                            \\
Tamil             & 38.66                                       & 69.84                            \\
Ukrainian         & 79.60                                       & 90.87                            \\
\textbf{Average}  & 81.10                                       & 87.98                            \\ \hline\hline
\end{tabular}
\caption{Compared ELAS scores on development set of fine-tuning single XLM-R embedding and ACE.}
\label{tab:single}
\end{table}

\begin{table*}[t!]
\small
\centering
\begin{tabular}{cccccccccc}
\hline\hline
& \multicolumn{9}{c}{Team Name}\\
\textbf{Language} & TGIF  & \textbf{Ours}  & ROBERTNLP & COMBO & UNIPI & DCU EPFL & GREW  & FASTPARSE & NUIG  \\ \hline
Arabic         & 81.23 & \textbf{82.26} & 81.58     & 76.39 & 77.17 & 71.01    & 71.13 & 53.74     & 0.00  \\
Bulgarian        & \textbf{93.63} & 92.52 & 93.16     & 86.67 & 90.84 & 92.44    & 88.83 & 78.73     & 78.45 \\
Czech        & \textbf{92.24} & 91.78 & 90.21     & 89.08 & 88.73 & 89.93    & 87.66 & 72.85     & 0.00  \\
Dutch        & \textbf{91.78} & 88.64 & 88.37     & 87.07 & 84.14 & 81.89    & 84.09 & 68.89     & 0.00  \\
English        & \textbf{88.19} & 87.27 & 87.88     & 84.09 & 87.11 & 85.70     & 85.49 & 73.00        & 65.40  \\
Estonian       & \textbf{88.38} & 86.66 & 86.55     & 84.02 & 81.27 & 84.35    & 78.19 & 60.05     & 54.03 \\
Finnish        & \textbf{91.75} & 90.81 & 91.01     & 87.28 & 89.62 & 89.02    & 85.20  & 57.71     & 0.00  \\
French        & \textbf{91.63} & 88.40  & 88.51     & 87.32 & 87.43 & 86.68    & 83.33 & 73.18     & 0.00  \\
Italian        & \textbf{93.31} & 92.88 & 93.28     & 90.40  & 91.81 & 92.41    & 90.98 & 78.32     & 0.00  \\
Latvian        & \textbf{90.23} & 89.17 & 88.82     & 84.57 & 83.01 & 86.96    & 77.45 & 66.43     & 56.67 \\
Lithuanian        & \textbf{86.06} & 80.87 & 80.76     & 79.75 & 71.31 & 78.04    & 74.62 & 48.27     & 59.13 \\
Polish        & \textbf{91.46} & 90.66 & 89.78     & 87.65 & 88.31 & 89.17    & 78.20  & 71.52     & 0.00  \\
Russian        & \textbf{94.01} & 93.59 & 92.64     & 90.73 & 90.90  & 92.83    & 90.56 & 78.56     & 66.33 \\
Slovak        & \textbf{94.96} & 90.25 & 89.66     & 87.04 & 86.05 & 89.59    & 86.92 & 64.28     & 67.45 \\
Swedish        & \textbf{89.90}  & 86.62 & 88.03     & 83.20  & 84.91 & 85.20     & 81.54 & 67.26     & 63.12 \\
Tamil        & \textbf{65.58} & 58.94 & 59.33     & 52.27 & 51.73 & 39.32    & 58.69 & 42.53     & 0.00  \\
Ukrainian        & \textbf{92.78} & 88.94 & 88.86     & 86.92 & 87.51 & 86.09    & 83.90  & 63.42     & 0.00  \\
Avg.      & \textbf{89.24} & 87.07 & 86.97     & 83.79 & 83.64 & 83.57    & 81.58 & 65.81     & 30.03 \\ \hline\hline
\end{tabular}
\caption{Official results of all systems. }
\label{tab:results}
\end{table*}

\begin{table*}[t!]
\small
\centering
\begin{tabular}{ccccc|cccc}
\hline
\hline
\multirow{2}{*}{\textbf{Language}} & \multicolumn{4}{c|}{\textbf{Stanza}}                                   & \multicolumn{4}{c}{\textbf{Trankit}}                                 \\\cline{2-9}
                  & \textbf{Tokens} & \textbf{Words} & \textbf{Sentences} & \textbf{ELAS} & \textbf{Tokens} & \textbf{Words} & \textbf{Sentences} & \textbf{ELAS} \\ \cline{2-9} 
Arabic            & 99.97           & 87.32          & 84.57              & 63.70          & 99.95           & 99.39          & 96.79              & 82.26         \\
Bulgarian         & 99.93           & 99.93          & 97.49              & 92.59         & 99.78           & 99.78          & 98.79              & 92.52         \\
Czech             & 99.92           & 99.92          & 95.03              & 91.50          & 99.93           & 99.92          & 97.56              & 91.78         \\
Dutch             & 99.94           & 99.94          & 82.32              & 89.62         & 99.00              & 99.00             & 83.48              & 88.64         \\
English           & 98.95           & 98.97          & 91.28              & 86.92         & 98.63           & 98.87          & 94.29              & 87.27         \\
Estonian          & 99.68           & 99.68          & 90.26              & 86.44         & 99.39           & 99.39          & 94.85              & 86.66         \\
Finnish           & 99.65           & 99.63          & 91.02              & 90.21         & 99.63           & 99.63          & 96.39              & 90.81         \\
French            & 99.60           & 99.39          & 95.61              & 87.60          & 99.76           & 99.75          & 97.23              & 88.40          \\
Italian           & 99.95           & 99.59          & 98.76              & 92.18         & 99.88           & 99.86          & 99.07              & 92.88         \\
Latvian           & 99.78           & 99.78          & 98.85              & 89.26         & 99.74           & 99.74          & 98.69              & 89.17         \\
Lithuanian        & 99.92           & 99.92          & 88.13              & 80.43         & 99.84           & 99.84          & 95.72              & 80.87         \\
Polish            & 99.51           & 99.54          & 98.26              & 89.58         & 99.47           & 99.92          & 99.05              & 90.66         \\
Russian           & 99.58           & 99.58          & 99.04              & 93.34         & 99.70            & 99.70           & 99.45              & 93.59         \\
Slovak            & 99.96           & 99.96          & 86.27              & 89.01         & 99.95           & 99.94          & 95.31              & 90.25         \\
Swedish           & 99.44           & 99.44          & 93.64              & 85.80          & 99.78           & 99.78          & 98.25              & 86.62         \\
Tamil             & 99.92           & 86.95          & 98.76              & 46.70          & 98.33           & 94.19          & 100.00                & 58.94         \\
Ukrainian         & 99.76           & 99.75          & 96.02              & 88.79         & 99.77           & 99.76          & 97.55              & 88.94         \\
\textbf{Average}  & 99.73           & 98.19          & 93.25              & 84.92         & 99.56           & 99.32          & 96.62              & 87.07         \\ \hline\hline
\end{tabular}
\caption{Comparison of different tokenization toolkits.}
\label{tab:token}
\end{table*}

\begin{table}[t!]
\begin{center}
\begin{tabular}{lr}
\hline \hline
\textbf{Hidden Layer} & \textbf{Hidden Sizes}\\ \hline
BiLSTM LSTM & 3*768 \\
Arc/Label & 500\\
Embedding/LSTM Dropouts & 33\%\\
Loss Interpolation ($\lambda$)& 0.025\\
Adam $\beta_1$ & 0.9\\
Adam $\beta_2$ & 0.9\\
Learning rate & $2e^{-3}$\\
LR decay & 0.5\\
\hline \hline
\end{tabular}
\end{center}
\caption{Hyper-parameters for our system. }
\label{tab:hyper_both}
\end{table}

\section{Settings and Results}
\subsection{Experimental Settings}
In training, we use the official development set as the development set. 
We tune the hyper-parameters on the development set and determine the hyper-parameter values according to the labeled F1 score (LF1) which is the evaluation metric used in SDP. LF1 measures the correctness of each arc-label pair. 
We use a batch size of 2000 tokens with the Adam \cite{kingma2014adam} optimizer. We set 30 steps of reinforcement learning, and the time of each reinforcement learning step depends on the size of data set. The hyper-parameters of our biaffine parser are shown in Table \ref{tab:hyper_both}, which are mostly adopted from previous work on dependency parsing. For the hyper-parameters of our ACE module, we follow the settings of \citet{wang2020automated}. We only use the tokenized words as the model input. For the sentence and word segmentation, we used the pretrained large model of \textit{trankit} \cite{nguyen2021trankit}. The embeddings we used in the ACE module for each language are shown in Table \ref{tab:embeddings}. For transformer-style embeddings, we only take the hidden states of the topmost layer and we only take the first piece subword representation as the multi-pieces word representation. 
We built our codes based on PyTorch \cite{paszke2019pytorch}, and trained the model for each language on a single Tesla V100 GPU.

\subsection{Main Results}
Table \ref{tab:single} shows the ELAS scores (defined as F1-score over the set of enhanced dependencies in the system output and the gold standard) on development set of biaffine parser with fine-tuning single XLM-R embedding and with our ACE module.
We can see that with ACE, the performance of most languages models is improved a lot.

Table \ref{tab:results} shows the results of official evaluations of all teams. We only show the ELAS in the results. We can see that our model gets the 1st on the Arabic language and gets the 2nd on averaged ELAS over 17 languages.

\subsection{Tokenization Performances of Different Toolkits}

In our experiments, we have tried two different tokenization toolkits. One is \textit{stanza} \citep{qi2020stanza} which is from Standford NLP Group, the others is \textit{trankit} \citep{nguyen2021trankit} which is a light-weight Transformer-based Python Toolkit for multilingual NLP. We use pretrained models of the two toolkits respectively. Furthermore, We train tokenization model of \textit{stanza} for each language. Both settings of \textit{stanza} are worse than \textit{trankit} on sentence segmentation score. Table \ref{tab:token} shows the sentences and words segmentation scores of \textit{stanza} trained on each language and pretrained \textit{trankit}. We see that although \textit{stanza} is better than \textit{trankit} on segmentation score of tokens, there is a huge performance gap on segmentation score of sentences between \textit{trankit} and \textit{stanza}. Therefore, the final ELAS on test set tokenized by \textit{trankit} is better than \textit{stanza}.

\section{Conclusion}
Our system is a parser with automated embeddings concatenation and a biaffine encoder. Empirical results show the effectiveness of 
ACE to enhanced universal dependencies. Our system ranks 2nd over 9 teams according to the official ELAS.

\bibliographystyle{acl_natbib}
\bibliography{anthology,acl2021}

\begin{thebibliography}{37}
\expandafter\ifx\csname natexlab\endcsname\relax\def\natexlab#1{#1}\fi

\bibitem[{Akbik et~al.(2018)Akbik, Blythe, and
  Vollgraf}]{akbik-etal-2018-contextual}
Alan Akbik, Duncan Blythe, and Roland Vollgraf. 2018.
\newblock \href {https://www.aclweb.org/anthology/C18-1139} {Contextual string
  embeddings for sequence labeling}.
\newblock In \emph{Proceedings of the 27th International Conference on
  Computational Linguistics}, pages 1638--1649, Santa Fe, New Mexico, USA.
  Association for Computational Linguistics.

\bibitem[{Arkhipov et~al.(2019)Arkhipov, Trofimova, Kuratov, and
  Sorokin}]{arkhipov-etal-2019-tuning}
Mikhail Arkhipov, Maria Trofimova, Yuri Kuratov, and Alexey Sorokin. 2019.
\newblock \href {https://doi.org/10.18653/v1/W19-3712} {Tuning multilingual
  transformers for language-specific named entity recognition}.
\newblock In \emph{Proceedings of the 7th Workshop on Balto-Slavic Natural
  Language Processing}, pages 89--93, Florence, Italy. Association for
  Computational Linguistics.

\bibitem[{Blinov and Avetisian(2020)}]{blinov-avetisian-2020-transformer}
Pavel Blinov and Manvel Avetisian. 2020.
\newblock \href {https://www.aclweb.org/anthology/2020.smm4h-1.17} {Transformer
  models for drug adverse effects detection from tweets}.
\newblock In \emph{Proceedings of the Fifth Social Media Mining for Health
  Applications Workshop {\&} Shared Task}, pages 110--112, Barcelona, Spain
  (Online). Association for Computational Linguistics.

\bibitem[{Bojanowski et~al.(2017)Bojanowski, Grave, Joulin, and
  Mikolov}]{bojanowski2017enriching}
Piotr Bojanowski, Edouard Grave, Armand Joulin, and Tomas Mikolov. 2017.
\newblock Enriching word vectors with subword information.
\newblock \emph{Transactions of the Association for Computational Linguistics},
  5:135--146.

\bibitem[{Bouma et~al.(2020)Bouma, Seddah, and Zeman}]{EUDparsingST:2020}
Gosse Bouma, Djam\'e Seddah, and Daniel Zeman. 2020.
\newblock {Overview of the IWPT 2020 Shared Task on Parsing into Enhanced
  Universal Dependencies}.
\newblock In \emph{{Proceedings of the 16th International Conference on Parsing
  Technologies and the IWPT 2020 Shared Task on Parsing into Enhanced Universal
  Dependencies}}, Seattle, US. Association for Computational Linguistics.

\bibitem[{Bouma et~al.(2021)Bouma, Seddah, and
  Zeman}]{bouma-etal-2021-overview}
Gosse Bouma, Djam\'e Seddah, and Daniel Zeman. 2021.
\newblock From {R}aw {T}ext to {E}nhanced {U}niversal {D}ependencies: the
  {P}arsing {S}hared {T}ask at {IWPT} 2021.
\newblock In \emph{Proceedings of the 17th International Conference on Parsing
  Technologies and the IWPT 2021 Shared Task on Parsing into Enhanced Universal
  Dependencies}, Online. Association for Computational Linguistics.

\bibitem[{Conneau et~al.(2020{\natexlab{a}})Conneau, Khandelwal, Goyal,
  Chaudhary, Wenzek, Guzm{\'a}n, Grave, Ott, Zettlemoyer, and
  Stoyanov}]{conneau2019unsupervised}
Alexis Conneau, Kartikay Khandelwal, Naman Goyal, Vishrav Chaudhary, Guillaume
  Wenzek, Francisco Guzm{\'a}n, Edouard Grave, Myle Ott, Luke Zettlemoyer, and
  Veselin Stoyanov. 2020{\natexlab{a}}.
\newblock \href {https://doi.org/10.18653/v1/2020.acl-main.747} {Unsupervised
  cross-lingual representation learning at scale}.
\newblock In \emph{Proceedings of the 58th Annual Meeting of the Association
  for Computational Linguistics}, pages 8440--8451, Online. Association for
  Computational Linguistics.

\bibitem[{Conneau et~al.(2020{\natexlab{b}})Conneau, Khandelwal, Goyal,
  Chaudhary, Wenzek, Guzm{\'a}n, Grave, Ott, Zettlemoyer, and
  Stoyanov}]{conneau-etal-2020-unsupervised}
Alexis Conneau, Kartikay Khandelwal, Naman Goyal, Vishrav Chaudhary, Guillaume
  Wenzek, Francisco Guzm{\'a}n, Edouard Grave, Myle Ott, Luke Zettlemoyer, and
  Veselin Stoyanov. 2020{\natexlab{b}}.
\newblock \href {https://doi.org/10.18653/v1/2020.acl-main.747} {Unsupervised
  cross-lingual representation learning at scale}.
\newblock In \emph{Proceedings of the 58th Annual Meeting of the Association
  for Computational Linguistics}, pages 8440--8451, Online. Association for
  Computational Linguistics.

\bibitem[{Delobelle et~al.(2020)Delobelle, Winters, and
  Berendt}]{delobelle-etal-2020-robbert}
Pieter Delobelle, Thomas Winters, and Bettina Berendt. 2020.
\newblock \href {https://doi.org/10.18653/v1/2020.findings-emnlp.292}
  {{R}ob{BERT}: a {D}utch {R}o{BERT}a-based {L}anguage {M}odel}.
\newblock In \emph{Findings of the Association for Computational Linguistics:
  EMNLP 2020}, pages 3255--3265, Online. Association for Computational
  Linguistics.

\bibitem[{Devlin et~al.(2018)Devlin, Chang, Lee, and
  Toutanova}]{devlin2018bert}
Jacob Devlin, Ming-Wei Chang, Kenton Lee, and Kristina Toutanova. 2018.
\newblock Bert: Pre-training of deep bidirectional transformers for language
  understanding.
\newblock \emph{arXiv preprint arXiv:1810.04805}.

\bibitem[{Devlin et~al.(2019)Devlin, Chang, Lee, and
  Toutanova}]{devlin-etal-2019-bert}
Jacob Devlin, Ming-Wei Chang, Kenton Lee, and Kristina Toutanova. 2019.
\newblock \href {https://doi.org/10.18653/v1/N19-1423} {{BERT}: Pre-training of
  deep bidirectional transformers for language understanding}.
\newblock In \emph{Proceedings of the 2019 Conference of the North {A}merican
  Chapter of the Association for Computational Linguistics: Human Language
  Technologies, Volume 1 (Long and Short Papers)}, pages 4171--4186,
  Minneapolis, Minnesota. Association for Computational Linguistics.

\bibitem[{Dozat and Manning(2018)}]{dozat2018simpler}
Timothy Dozat and Christopher~D Manning. 2018.
\newblock Simpler but more accurate semantic dependency parsing.
\newblock \emph{arXiv preprint arXiv:1807.01396}.

\bibitem[{Kingma and Ba(2015)}]{kingma2014adam}
Diederik~P Kingma and Jimmy Ba. 2015.
\newblock Adam: A method for stochastic optimization.
\newblock In \emph{International Conference on Learning Representations}.

\bibitem[{Liu et~al.(2019)Liu, Ott, Goyal, Du, Joshi, Chen, Levy, Lewis,
  Zettlemoyer, and Stoyanov}]{liu2019roberta}
Yinhan Liu, Myle Ott, Naman Goyal, Jingfei Du, Mandar Joshi, Danqi Chen, Omer
  Levy, Mike Lewis, Luke Zettlemoyer, and Veselin Stoyanov. 2019.
\newblock Roberta: A robustly optimized bert pretraining approach.
\newblock \emph{arXiv preprint arXiv:1907.11692}.

\bibitem[{Malmsten et~al.(2020)Malmsten, Börjeson, and
  Haffenden}]{swedish-bert}
Martin Malmsten, Love Börjeson, and Chris Haffenden. 2020.
\newblock \href {http://arxiv.org/abs/2007.01658} {Playing with words at the
  national library of sweden -- making a swedish bert}.

\bibitem[{Martin et~al.(2020)Martin, Muller, Ortiz~Su{\'a}rez, Dupont, Romary,
  de~la Clergerie, Seddah, and Sagot}]{martin-etal-2020-camembert}
Louis Martin, Benjamin Muller, Pedro~Javier Ortiz~Su{\'a}rez, Yoann Dupont,
  Laurent Romary, {\'E}ric de~la Clergerie, Djam{\'e} Seddah, and Beno{\^\i}t
  Sagot. 2020.
\newblock \href {https://doi.org/10.18653/v1/2020.acl-main.645} {{C}amem{BERT}:
  a tasty {F}rench language model}.
\newblock In \emph{Proceedings of the 58th Annual Meeting of the Association
  for Computational Linguistics}, pages 7203--7219, Online. Association for
  Computational Linguistics.

\bibitem[{McDonald et~al.(2005)McDonald, Crammer, and
  Pereira}]{mcdonald2005online}
Ryan McDonald, Koby Crammer, and Fernando Pereira. 2005.
\newblock Online large-margin training of dependency parsers.
\newblock In \emph{Proceedings of the 43rd Annual Meeting of the Association
  for Computational Linguistics (ACL’05)}, pages 91--98.

\bibitem[{Mikolov et~al.(2013)Mikolov, Sutskever, Chen, Corrado, and
  Dean}]{mikolov2013distributed}
Tomas Mikolov, Ilya Sutskever, Kai Chen, Greg~S Corrado, and Jeff Dean. 2013.
\newblock Distributed representations of words and phrases and their
  compositionality.
\newblock In \emph{Advances in neural information processing systems}, pages
  3111--3119.

\bibitem[{Nguyen et~al.(2021)Nguyen, Lai, Veyseh, and
  Nguyen}]{nguyen2021trankit}
Minh~Van Nguyen, Viet Lai, Amir Pouran~Ben Veyseh, and Thien~Huu Nguyen. 2021.
\newblock Trankit: A light-weight transformer-based toolkit for multilingual
  natural language processing.
\newblock In \emph{Proceedings of the 16th Conference of the European Chapter
  of the Association for Computational Linguistics: System Demonstrations}.

\bibitem[{Nivre et~al.(2016)Nivre, de~Marneffe, Ginter, Goldberg, Haji{\v{c}},
  Manning, McDonald, Petrov, Pyysalo, Silveira, Tsarfaty, and
  Zeman}]{nivre-etal-2016-universal}
Joakim Nivre, Marie-Catherine de~Marneffe, Filip Ginter, Yoav Goldberg, Jan
  Haji{\v{c}}, Christopher~D. Manning, Ryan McDonald, Slav Petrov, Sampo
  Pyysalo, Natalia Silveira, Reut Tsarfaty, and Daniel Zeman. 2016.
\newblock \href {https://www.aclweb.org/anthology/L16-1262} {{U}niversal
  {D}ependencies v1: A multilingual treebank collection}.
\newblock In \emph{Proceedings of the Tenth International Conference on
  Language Resources and Evaluation ({LREC}'16)}, pages 1659--1666,
  Portoro{\v{z}}, Slovenia. European Language Resources Association (ELRA).

\bibitem[{Oepen et~al.(2015)Oepen, Kuhlmann, Miyao, Zeman, Cinkov{\'a},
  Flickinger, Hajic, and Uresova}]{oepen2015semeval}
Stephan Oepen, Marco Kuhlmann, Yusuke Miyao, Daniel Zeman, Silvie Cinkov{\'a},
  Dan Flickinger, Jan Hajic, and Zdenka Uresova. 2015.
\newblock Semeval 2015 task 18: Broad-coverage semantic dependency parsing.
\newblock In \emph{Proceedings of the 9th International Workshop on Semantic
  Evaluation (SemEval 2015)}, pages 915--926.

\bibitem[{Paszke et~al.(2019)Paszke, Gross, Massa, Lerer, Bradbury, Chanan,
  Killeen, Lin, Gimelshein, Antiga et~al.}]{paszke2019pytorch}
Adam Paszke, Sam Gross, Francisco Massa, Adam Lerer, James Bradbury, Gregory
  Chanan, Trevor Killeen, Zeming Lin, Natalia Gimelshein, Luca Antiga, et~al.
  2019.
\newblock Pytorch: An imperative style, high-performance deep learning library.
\newblock In \emph{Advances in Neural Information Processing Systems}, pages
  8024--8035.

\bibitem[{Pennington et~al.(2014)Pennington, Socher, and
  Manning}]{pennington2014glove}
Jeffrey Pennington, Richard Socher, and Christopher Manning. 2014.
\newblock Glove: Global vectors for word representation.
\newblock In \emph{Proceedings of the 2014 conference on empirical methods in
  natural language processing (EMNLP)}, pages 1532--1543.

\bibitem[{Peters et~al.(2018)Peters, Neumann, Iyyer, Gardner, Clark, Lee, and
  Zettlemoyer}]{peters-etal-2018-deep}
Matthew Peters, Mark Neumann, Mohit Iyyer, Matt Gardner, Christopher Clark,
  Kenton Lee, and Luke Zettlemoyer. 2018.
\newblock \href {https://doi.org/10.18653/v1/N18-1202} {Deep contextualized
  word representations}.
\newblock In \emph{Proceedings of the 2018 Conference of the North {A}merican
  Chapter of the Association for Computational Linguistics: Human Language
  Technologies, Volume 1 (Long Papers)}, pages 2227--2237, New Orleans,
  Louisiana. Association for Computational Linguistics.

\bibitem[{Qi et~al.(2020)Qi, Zhang, Zhang, Bolton, and Manning}]{qi2020stanza}
Peng Qi, Yuhao Zhang, Yuhui Zhang, Jason Bolton, and Christopher~D Manning.
  2020.
\newblock Stanza: A python natural language processing toolkit for many human
  languages.
\newblock \emph{arXiv preprint arXiv:2003.07082}.

\bibitem[{Safaya et~al.(2020)Safaya, Abdullatif, and
  Yuret}]{safaya-etal-2020-kuisail}
Ali Safaya, Moutasem Abdullatif, and Deniz Yuret. 2020.
\newblock \href {https://www.aclweb.org/anthology/2020.semeval-1.271}
  {{KUISAIL} at {S}em{E}val-2020 task 12: {BERT}-{CNN} for offensive speech
  identification in social media}.
\newblock In \emph{Proceedings of the Fourteenth Workshop on Semantic
  Evaluation}, pages 2054--2059, Barcelona (online). International Committee
  for Computational Linguistics.

\bibitem[{Santos and Zadrozny(2014)}]{santos2014learning}
Cicero~D Santos and Bianca Zadrozny. 2014.
\newblock Learning character-level representations for part-of-speech tagging.
\newblock In \emph{Proceedings of the 31st international conference on machine
  learning (ICML-14)}, pages 1818--1826.

\bibitem[{Strakov{\'a} et~al.(2019)Strakov{\'a}, Straka, and
  Hajic}]{strakova-etal-2019-neural}
Jana Strakov{\'a}, Milan Straka, and Jan Hajic. 2019.
\newblock \href {https://doi.org/10.18653/v1/P19-1527} {Neural architectures
  for nested {NER} through linearization}.
\newblock In \emph{Proceedings of the 57th Annual Meeting of the Association
  for Computational Linguistics}, pages 5326--5331, Florence, Italy.
  Association for Computational Linguistics.

\bibitem[{Ul{\v c}ar and Robnik-{\v S}ikonja(2020)}]{20.500.11821/42}
Matej Ul{\v c}ar and Marko Robnik-{\v S}ikonja. 2020.
\newblock \href {http://hdl.handle.net/20.500.11821/42} {{LitLat} {BERT}}.
\newblock {CLARIN}-{LT} digital library in the Republic of Lithuania.

\bibitem[{Virtanen et~al.(2019)Virtanen, Kanerva, Ilo, Luoma, Luotolahti,
  Salakoski, Ginter, and Pyysalo}]{DBLP:journals/corr/abs-1912-07076}
Antti Virtanen, Jenna Kanerva, Rami Ilo, Jouni Luoma, Juhani Luotolahti, Tapio
  Salakoski, Filip Ginter, and Sampo Pyysalo. 2019.
\newblock \href {http://arxiv.org/abs/1912.07076} {Multilingual is not enough:
  {BERT} for finnish}.
\newblock \emph{CoRR}, abs/1912.07076.

\bibitem[{Wang et~al.(2019)Wang, Huang, and Tu}]{wang-etal-2019-second}
Xinyu Wang, Jingxian Huang, and Kewei Tu. 2019.
\newblock \href {https://doi.org/10.18653/v1/P19-1454} {Second-order semantic
  dependency parsing with end-to-end neural networks}.
\newblock In \emph{Proceedings of the 57th Annual Meeting of the Association
  for Computational Linguistics}, pages 4609--4618, Florence, Italy.
  Association for Computational Linguistics.

\bibitem[{Wang et~al.(2021)Wang, Jiang, Bach, Wang, Huang, Huang, and
  Tu}]{wang2020automated}
Xinyu Wang, Yong Jiang, Nguyen Bach, Tao Wang, Zhongqiang Huang, Fei Huang, and
  Kewei Tu. 2021.
\newblock {{Automated Concatenation of Embeddings for Structured Prediction}}.
\newblock In \emph{{the Joint Conference of the 59th Annual Meeting of the
  Association for Computational Linguistics and the 11th International Joint
  Conference on Natural Language Processing (\textbf{ACL-IJCNLP 2021})}}.
  Association for Computational Linguistics.

\bibitem[{Wang et~al.(2020{\natexlab{a}})Wang, Jiang, Bach, Wang, Zhongqiang,
  Huang, and Tu}]{wang-etal-2020-more}
Xinyu Wang, Yong Jiang, Nguyen Bach, Tao Wang, Huang Zhongqiang, Fei Huang, and
  Kewei Tu. 2020{\natexlab{a}}.
\newblock More embeddings, better sequence labelers?
\newblock In \emph{Findings of EMNLP}, Online.

\bibitem[{Wang et~al.(2020{\natexlab{b}})Wang, Jiang, and
  Tu}]{wang-etal-2020-enhanced}
Xinyu Wang, Yong Jiang, and Kewei Tu. 2020{\natexlab{b}}.
\newblock \href {https://doi.org/10.18653/v1/2020.iwpt-1.22} {Enhanced
  {U}niversal {D}ependency parsing with second-order inference and mixture of
  training data}.
\newblock In \emph{Proceedings of the 16th International Conference on Parsing
  Technologies and the IWPT 2020 Shared Task on Parsing into Enhanced Universal
  Dependencies}, pages 215--220, Online. Association for Computational
  Linguistics.

\bibitem[{Wang and Tu(2020)}]{wang-tu-2020-second}
Xinyu Wang and Kewei Tu. 2020.
\newblock \href {https://www.aclweb.org/anthology/2020.aacl-main.12}
  {Second-order neural dependency parsing with message passing and end-to-end
  training}.
\newblock In \emph{Proceedings of the 1st Conference of the Asia-Pacific
  Chapter of the Association for Computational Linguistics and the 10th
  International Joint Conference on Natural Language Processing}, pages 93--99,
  Suzhou, China. Association for Computational Linguistics.

\bibitem[{Williams(1992)}]{williams1992simple}
Ronald~J Williams. 1992.
\newblock Simple statistical gradient-following algorithms for connectionist
  reinforcement learning.
\newblock \emph{Machine learning}, 8(3-4):229--256.

\bibitem[{Yang et~al.(2019)Yang, Dai, Yang, Carbonell, Salakhutdinov, and
  Le}]{yang2019xlnet}
Zhilin Yang, Zihang Dai, Yiming Yang, Jaime Carbonell, Russ~R Salakhutdinov,
  and Quoc~V Le. 2019.
\newblock Xlnet: Generalized autoregressive pretraining for language
  understanding.
\newblock In \emph{Advances in neural information processing systems}, pages
  5753--5763.

\end{thebibliography}


\end{document}